\documentclass[letterpaper,twocolumn,fleqn]{article} 

\usepackage{ist}
\usepackage{amsmath}
\newcommand{\argmax}{\arg\!\max}
\title{Blockwise Based Detection of Local Defects$^*$}

\author{ Xiaoyu Xiang$^a$, Renee Jessome$^b$, Eric Maggard$^b$, Yousun Bang$^c$, Minki Cho$^c$, Jan Allebach$^a$ \newline
$^a$School of Electrical and Computer Engineering, Purdue University, West Lafayette, IN 47907, USA \newline
$^b$HP Inc. Boise, ID 83714, USA\newline
$^c$HPPK (HP Printing Korea), Suwon City, Gyeonggi, 16677, Republic of Korea
}

\date{} 

\begin{document} 
\maketitle 
\thispagestyle{empty} 
\begin{abstract}
Print quality is an important criterion for a printer's performance. The detection, classification, and assessment of printing defects can reflect the printer's working status and help to locate mechanical problems inside. To handle all these questions, an efficient algorithm is needed to replace the traditionally visual checking method. In this paper, we focus on pages with local defects including gray spots and solid spots. We propose a coarse-to-fine method to detect local defects in a block-wise manner, and aggregate the blockwise attributes to generate the feature vector of the whole test page for a further ranking task. In the detection part, we first select candidate regions by thresholding a single feature. Then more detailed features of candidate blocks are calculated and sent to a decision tree that is previously trained on our training dataset. The final result is given by the decision tree model to control the false alarm rate while maintaining the required miss rate. Our algorithm is proved to be effective in detecting and classifying local defects compared with previous methods.{\let\thefootnote\relax\footnote{{* This work was supported by HP Inc, Boise, ID 83714, USA}}}
\end{abstract}

\section{Introduction}
\label{sec:intro}

Laser electrophotographic (EP) printers have been widely used in past decades. As one of the most important criteria in evaluating the performance of a printer, print quality is not only of concern to customers, but also designers of printers. For the reason that print quality can reflect the current working status and reveal hidden mechanical problems inside of a printer, the assessment of print quality has continued to be an important topic in printer-related research.

The traditional way of print quality diagnosis relies on manual examination of a printed page, which is specially designed for the testing purpose. The assessment work that is usually conducted by well-trained experts, includes marking exact areas with local defects and rating the overall page. Each test page can be rated as ``A" ``B" ``C" or ``D" four ranks, in which ``A" and ``B" mean the page passes the print quality assessment, while ``C" and ``D" mean the page fails the assessment. However, given the large number of pages to be evaluated, manually examining all pages is too costly and time-consuming. To solve this problem, a print defect detection algorithm is highly desired for building a smart print quality diagnosis system. 

\begin{figure}[!hb]
  \includegraphics[width=1.0\columnwidth]{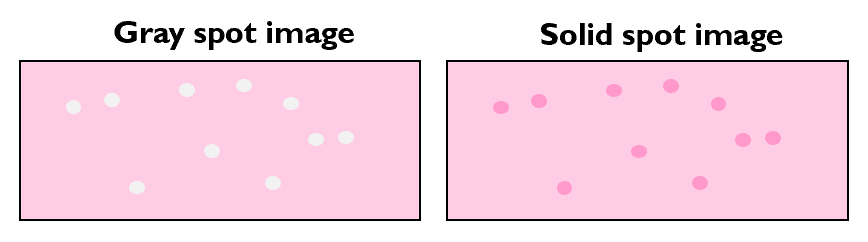}
  \caption{Comparison of simulated gray spots and solid spots.}
  \label{Figure:gs_ss}
\end{figure}

The local defect is one of the print defects of most critical concern. Typical local defects include gray spots and solid spots. The gray spot (also called carrier spot) is a phenomenon of low density around the agglomerates, which usually happens when the toner transfer from the Organic Photoconductor (OPC) to the Intermediate Transfer Belt (ITB) is blocked by some developed carriers or toner agglomerates on the OPC. Thus, a poor transfer of toners occurs around the agglomerates. An obvious visual feature of gray spots is that their color is lighter than that of the surrounding content (as shown in Figure \ref{Figure:gs_ss}). 

The solid spot is another type of local defect. Different from gray spots, the color of solid spots is usually darker than nearby contents (as shown in Figure \ref{Figure:gs_ss})). This phenomenon of high density around is due to the agglomerates of toner or carriers. The solid spot is a defect that occurs during the retransfer process, and is often observed in halftone patterns. Generally, the cause of solid spots is that toner retransfer from the ITB to the OPC is blocked by some developed carriers or toner agglomerates on the ITB. When the after-image transferred on the ITB moves to the next Pod (T1 or T2), an air gap occurs by the carrier on non-image area or contamination. Since the retransfer quantity is lower in the air gap area, the area appears as a solid spot.

There are some previous works on automatic detection of print defects. Jing et al. \cite{jing2013general} borrowed a metric from the image quality area to print defect assessment. Ju et al. \cite{ju2015autonomous}, Yan et al. \cite{yan2015autonomous} and Xiao et al. \cite{xiao2018detection} proposed new algorithms to predict the visibility of fading defects. Zhang et al. \cite{zhang2015estimation}\cite{zhang2013assessment} modeled periodic and aperiodic bands, and applied a histogram-based method to detect them. Nguyen et al. \cite{nguyen2015controlling}\cite{nguyen2017feature}\cite{nguyen2014perceptual} designed a complete framework for print defect prediction based on defined intra-block and inter-block features for local and global characteristics, respectively. For local defects, Wang et al. \cite{wang2016local} developed an algorithm to detect them and predict overall print quality with a trained support vector machine (SVM). In previous papers, we have a limited understanding of the cause, type, and severity of local defects. Different from streaks \cite{zhang2019streak} or banding \cite{huang2019banding}, we still lack a model describing common local defects. 

In this paper, we develop a blockwise algorithm to detect and characterize local defects. This method involves a coarse-to-fine strategy in defect areas detection: first select possible regions by simple thresholding, and then apply a decision tree to exclude false alarms. In addition, our algorithm can classify different local defects according to their perceptual attributes including size, brightness, and other aspects. 

\section{Methodology}
\begin{figure}[!hb]
  \includegraphics[width=\columnwidth]{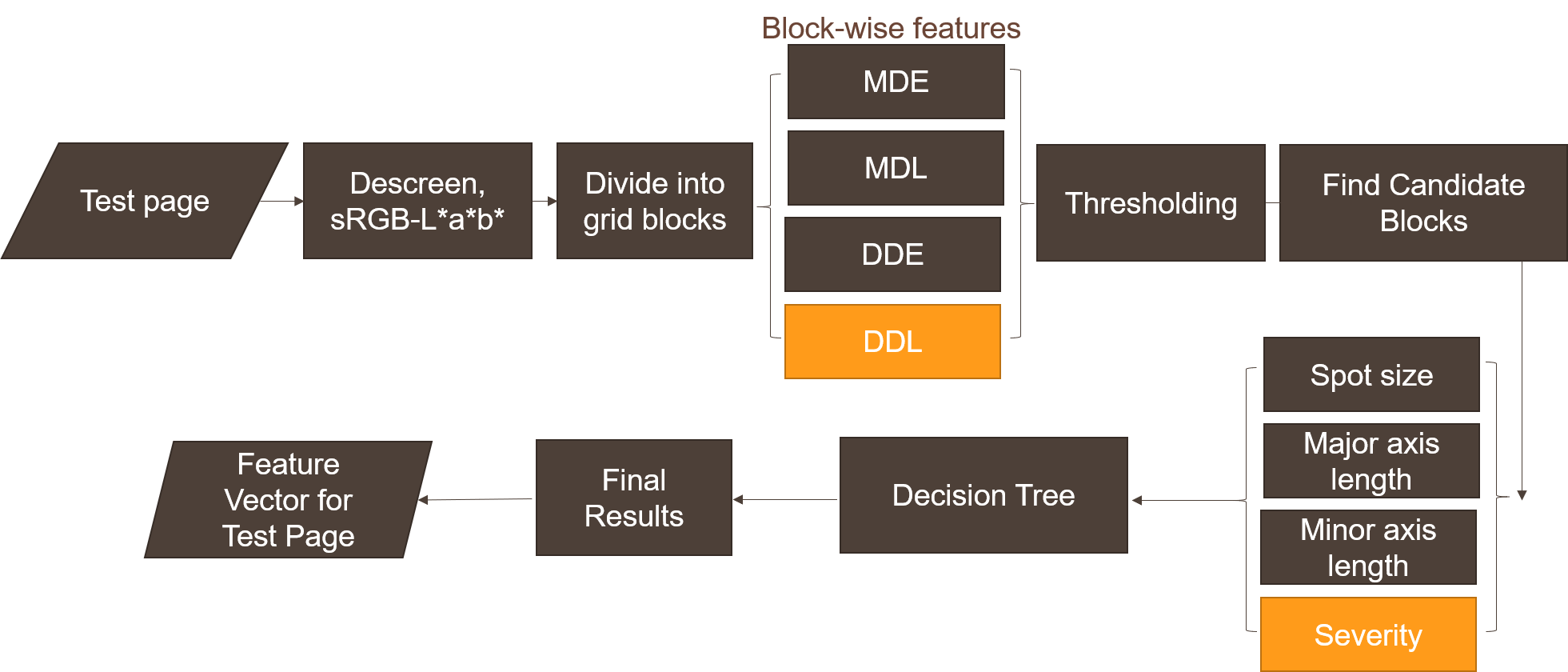}
  \caption{The pipeline of our method.}
  \label{Figure:pipeline}
\end{figure}

The overall workflow of our method is shown in Figure \ref{Figure:pipeline}. The detection of local defects can be roughly divided into two stages: finding candidate areas, and verify the defect features inside each candidate block to give the final results. The two-stage detection pipeline can greatly reduce the runtime while ensuring the required miss rate. 

\subsection{Preprocessing Test Page}

\subsubsection{Masking}

\begin{figure}[!hb]
  \includegraphics[width=1.0\columnwidth]{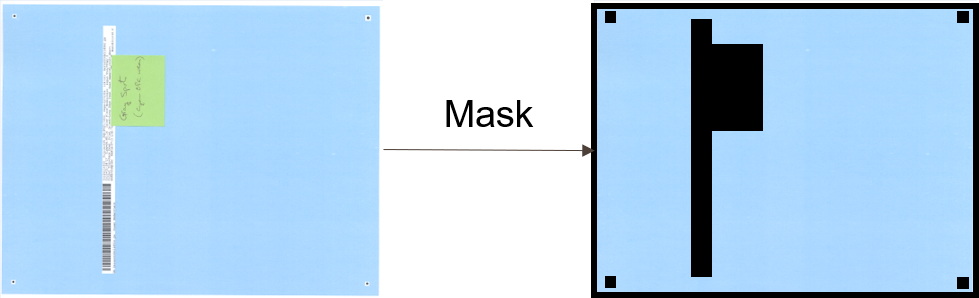}
  \caption{Test page sample (Original Page and masked page).}
  \label{Figure:mask}
\end{figure}

The test pages are letter-size pages with at least one constant-tint area that is printed with one solid color (cyan, magenta, red, etc). These test pages are scanned at 600 dpi and stored in Portable Network Graphs (PNG format) that includes an alpha channel as a mask, which is a binary channel where 0 (black) stands for the masked part while 1 (white) is for the content. The mask channel is used to tell our algorithm which part of the test page should be processed. Because besides those constant-tone areas that our defect detection algorithm focuses on, each test page also has some other contents such as fiducial marks, and a barcode that records metadata about the master file and the original printer. Also, the unprinted areas, e.g.,  white edge, should also be excluded from our area of interest. Figure \ref{Figure:mask} shows an original image, and the masked image of one test page.

\subsubsection{Descreening}

\begin{figure}[!hb]
  \includegraphics[width=1.0\columnwidth]{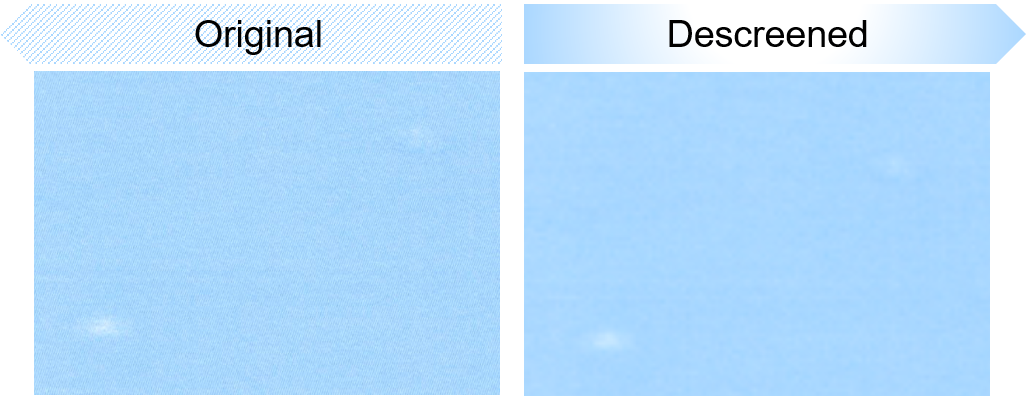}
  \caption{Comparison of the original page and descreened page.}
  \label{Figure:descreen}
\end{figure}

Since all the pages are printed as halftones, high-resolution scanning would show the halftone patterns in our test pages. This might cause abnormal false alarms for local defect detection. In order to remove the halftone patterns without ruining the local defects, we apply a $12\times 12$ Gaussian filter with a standard deviation of 2. Figure \ref{Figure:descreen} shows a test area before and after descreening. It is clear to see that the processed region is smoothed with no visible halftone patterns while maintaining the gray spots in which we interested.

\subsubsection{Color Space Conversion}
After descreening the half-toned pages, we need to transfer the pixels from the RGB color space to the CIE L*a*b* color space, where $L*$ is the lightness and $a*$ and $b*$ are the green-red and blue-yellow color components, respectively. Compared with the RGB color space, L*a*b* is designed to be perceptually uniform according to human color vision \cite{fairchild2013color}. Thus, the L*a*b* color space is widely used in color comparisons.

\subsection{Blockwise Detection of Local Defects}
\subsubsection{Select Candidate Blocks}
\begin{figure}[!hb]
\centering
  \includegraphics[width=0.35\columnwidth]{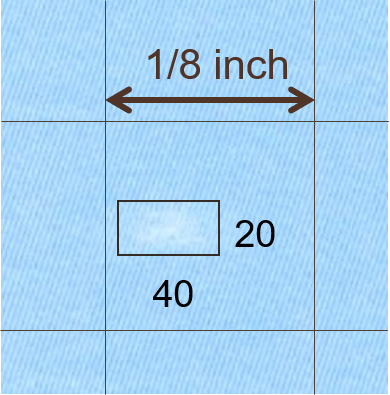}
  \caption{Choosing block size based on defect size.}
  \label{Figure:size}
\end{figure}

In order to detect local defects, we first divide the big region into relatively small blocks and detect local defects in each block. The advantage of blockwise detection is to lower memory consumption. We choose $75 \times 75$ as block size according to the scale of common local defects (see Figure \ref{Figure:size}), so that a block can be big enough contain a complete local defect, but not too big that contain multiple defects that might interrupt following computations. 

\begin{figure}[!hb]
  \begin{tabular}{@{}c@{}}
    \includegraphics[width=\linewidth]{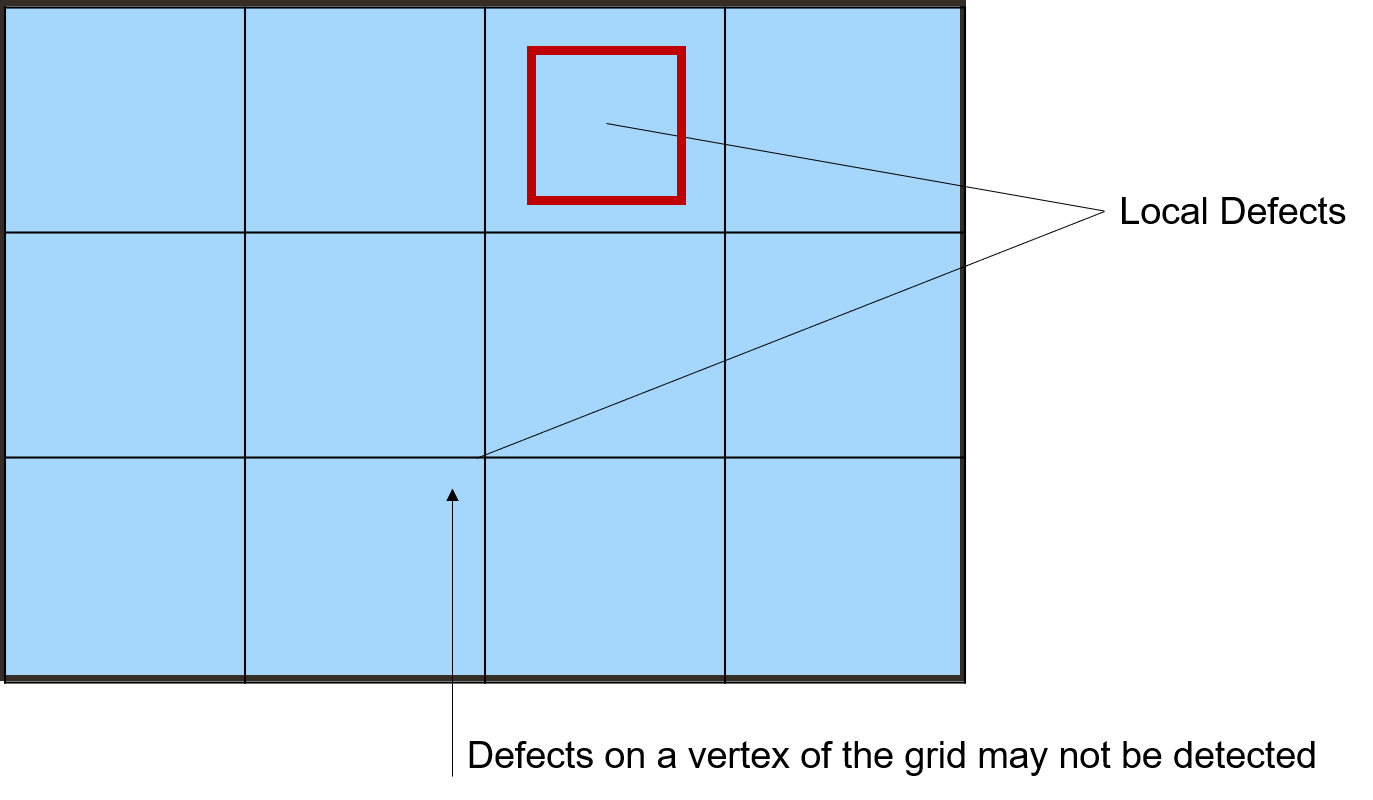} \\[\abovecaptionskip]
    \small (a) Initial grid
  \end{tabular}

  \begin{tabular}{@{}c@{}}
    \includegraphics[width=\linewidth]{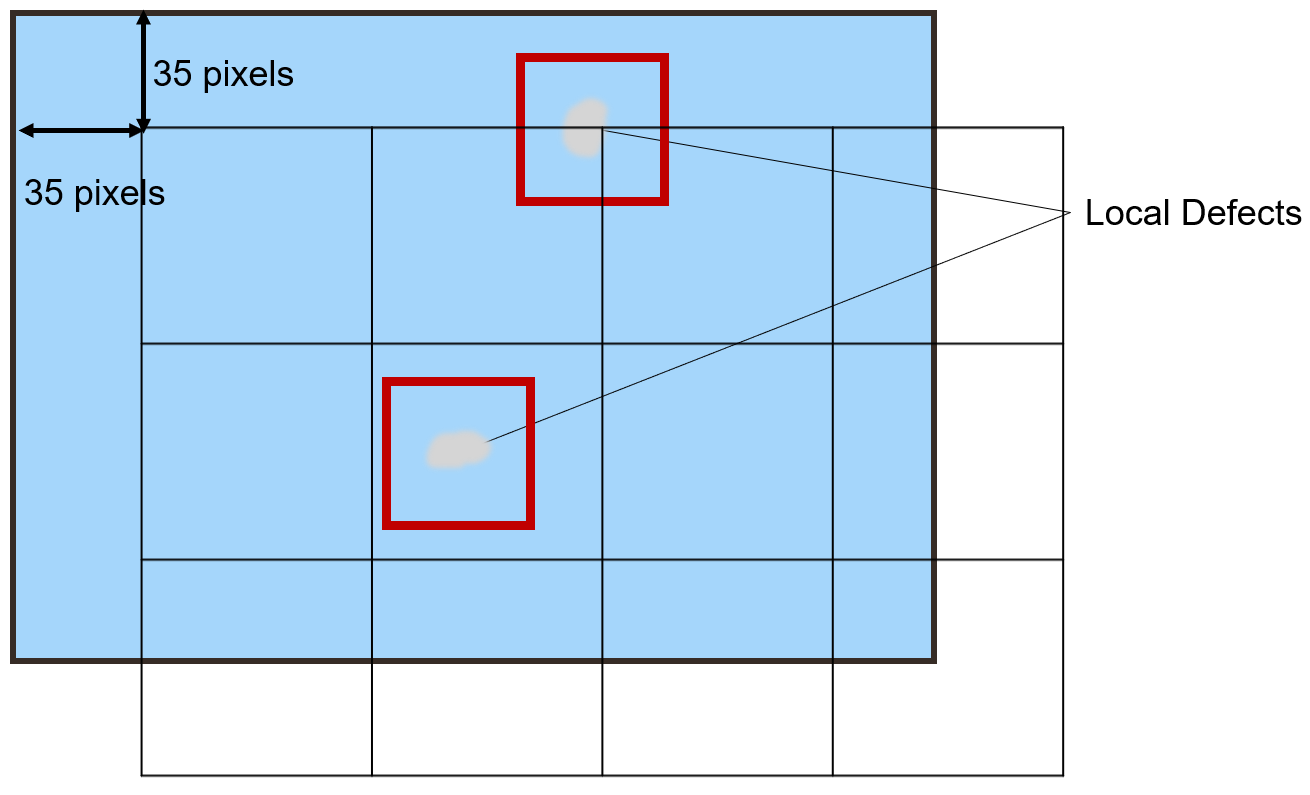} \\[\abovecaptionskip]
    \small (b) Move the grid by 35 pixels in both directions
  \end{tabular}
  
    \vspace{\floatsep}

  \caption{Move the grid to detect defects in all possible positions. After performing the detection twice, we combine the detected blocks as the Region of Interest (ROI)}\label{fig:2steps}
\end{figure}

Since the local defects are randomly located on our test page, it is very likely that a local defect falls on the boundary or even a vertex of the grid. If such a case happens, these local defects might be hard to detect. So it is necessary to search a second time for the missing defects. In the second detection, we move the grid by 35 pixels in both $x$ and $y$ directions from its initial location. Figure \ref{fig:2steps} shows the difference between the two grids. The local defects that cannot be detected in the first time would fall in the middle of the block. So we just need to combine the two detection results to get all local defects. However, there can be overlaps between two detection results. The connected components algorithm is applied to count the local defects accurately. The combined output is our initial estimation of areas with local defects, or the Region of Interest (ROI). 

For each block, we use the metrics of graininess on the local scale that were first defined by Nguyen et al. \cite{nguyen2014perceptual}. These metrics have proved to be effective in the following works \cite{nguyen2015controlling}\cite{wang2016local}. In general, Nguyen et al. quantized the intra-block fluctuation by computing the RMS (root mean square) difference from the mean for each pixel in a block. 

In the pre-processing step, the input page is converted to L*a*b* color space. So for each block, we can compute the average L*, a*, and b* according to every pixel's L*a*b* value. Then we can compute the difference between the block average and a pixel $i$ inside of block $j$:

\begin{equation}
    \Delta E_{ij} = \sqrt{(L_{ij}^* - L_{block_j}^*)^2 + (a_{ij}^* - a_{block_j}^*)^2 + (b_{ij}^* - b_{block_j}^*)^2}
\end{equation}

where $L_{block_j}^*$, $a_{block_j}^*$, $b_{block_j}^*$ denote average values within the block $j$, and $L_{ij}^*$, $a_{ij}^*$, $b_{ij}^*$ stand for the pixel values. After finishing the calculations all the pixels, the mean $\Delta E$ (MDE) for a block $j$ can be computed:

\begin{equation}
    MDE_j = \frac{1}{75^2} \sum_{i=1}^{75^2} \Delta E_{ij}
\end{equation}

The standard deviation of $\Delta E$ (DDE) is given by:

\begin{equation}
    DDE_j = \sqrt{\frac{1}{75^2-1} \sum_{i=1}^{75^2}(\Delta E_{ij} - MDE_j)^2}
\end{equation}

In a similar manner, we can define $L*$ related metrics $\Delta L$, $MDL$, and $DDL$:

\begin{equation}
\Delta L_{ij}^* = |L_{ij}^* - L_{block_j}^*|    
\end{equation}

\begin{equation}
    MDL_j = \frac{1}{75^2} \sum_{i=1}^{75^2} \Delta L_{ij}^*
\end{equation}

\begin{equation}
    DDL_j = \sqrt{\frac{1}{75^2-1} \sum_{i=1}^{75^2}(\Delta L_{ij}^* - MDL_j)^2}
\end{equation}

\begin{figure}
  \begin{tabular}{@{}c@{}}
    \includegraphics[width=\linewidth]{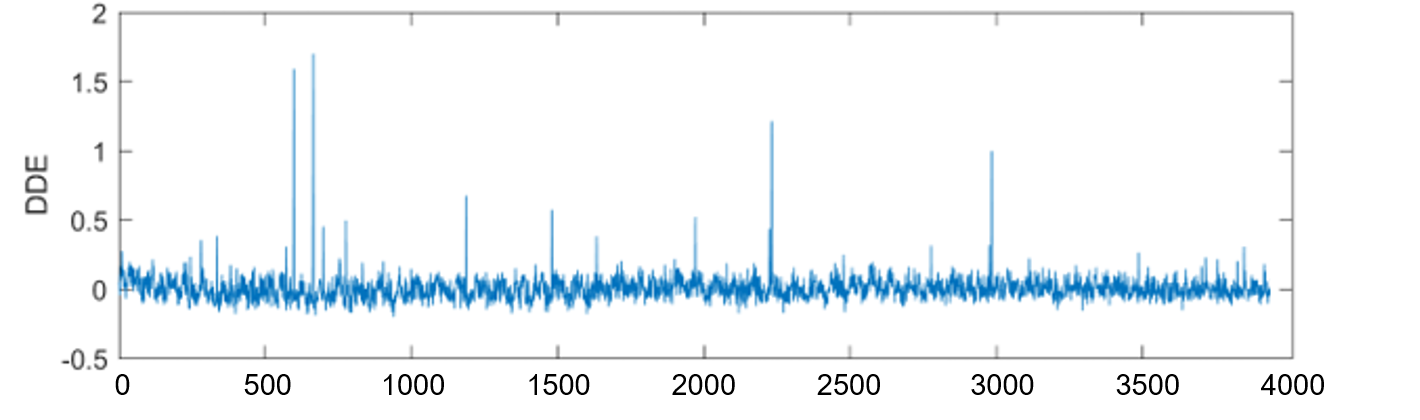} \\[\abovecaptionskip]
    \small (a) Plot DDE of each block
  \end{tabular}

  \begin{tabular}{@{}c@{}}
    \includegraphics[width=\linewidth]{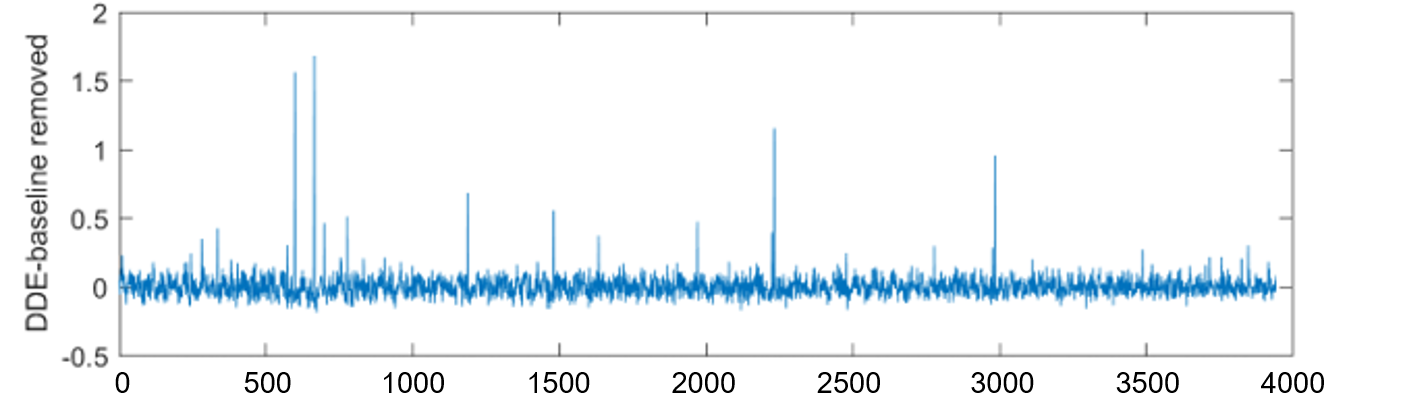} \\[\abovecaptionskip]
    \small (b) Remove baseline
  \end{tabular}
  
  \begin{tabular}{@{}c@{}}
    \includegraphics[width=\linewidth]{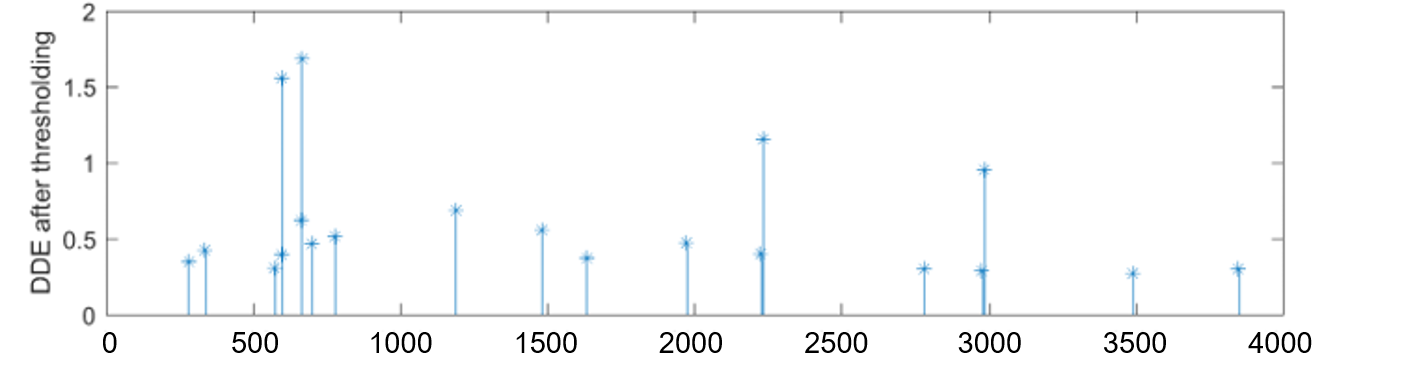} \\[\abovecaptionskip]
    \small (c) Select candidate blocks according to DDE
  \end{tabular}
  \vspace{\floatsep}
  \caption{Select candidate blocks according to their DDE}\label{fig:3steps}
\end{figure}

Repeat the calculations above until we go over all blocks. Higher $DDE$ is related to a block with more fluctuations, thus it is more likely to have local defects, We take the $DDE$ as the metric to get the candidate ROI. As shown in Figure \ref{fig:3steps}, first we remove the baseline of DDE to make our algorithm less sensitive to local noise. Then we need to filter out the blocks with less fluctuation by thresholding. The peaks remaining in the last graph identify the blocks that comprise the ROI.

\subsubsection{Features of Local Defects}
In this section, we will introduce how to find the visible defects in the candidate blocks. Usually, local defects are small spots that are lighter or darker than the background. So we adopted valley-emphasis algorithm to mark the distinctive pixels in a candidate block. Ostu's algorithm \cite{otsu1979threshold} is a popular method, where the preferred threshold $t$ is chosen automatically by maximizing the between-class variance:

\begin{equation}
    t^* = \argmax_{0\leq t\leq L} [\omega_1(t)\mu_1(t)^2 + \omega^2(t)\mu_2(t)^2]
\end{equation}

where $t$ denotes the input (e.g. $\Delta E$, $L^*$) value, $L$ is the number of distinct gray levels, $\omega_1$, $\omega_2$ are percentages of pixels that belong to the background and defects, respectively, and $\mu_1$ and $\mu_2$ are the average values of background and defect pixels.

Ng et al. \cite{ng2006automatic} proposed a new form of valley-emphasis algorithm based on Otsu's method by adding a new term in the maximization to emphasize the "valley" in the histogram. It is based on the assumption that the correct threshold should be located at the "valley" of histogram so that the foreground and background can be separated properly. Ng et al's modification can be expressed as:

\begin{equation}
    t^* = \argmax_{0\leq t \leq L} [(1-p(t))(\omega_1(t)\mu_1(t)^2 + \omega^2(t)\mu_2(t)^2)]
\end{equation}

The additional term $p(t)$ is the percentage of pixels at level $t$. It serves as a weighting term, so that the lower the percentage is, the higher the weight is.

\begin{figure}[!hb]
\centering
  \includegraphics[width=\columnwidth]{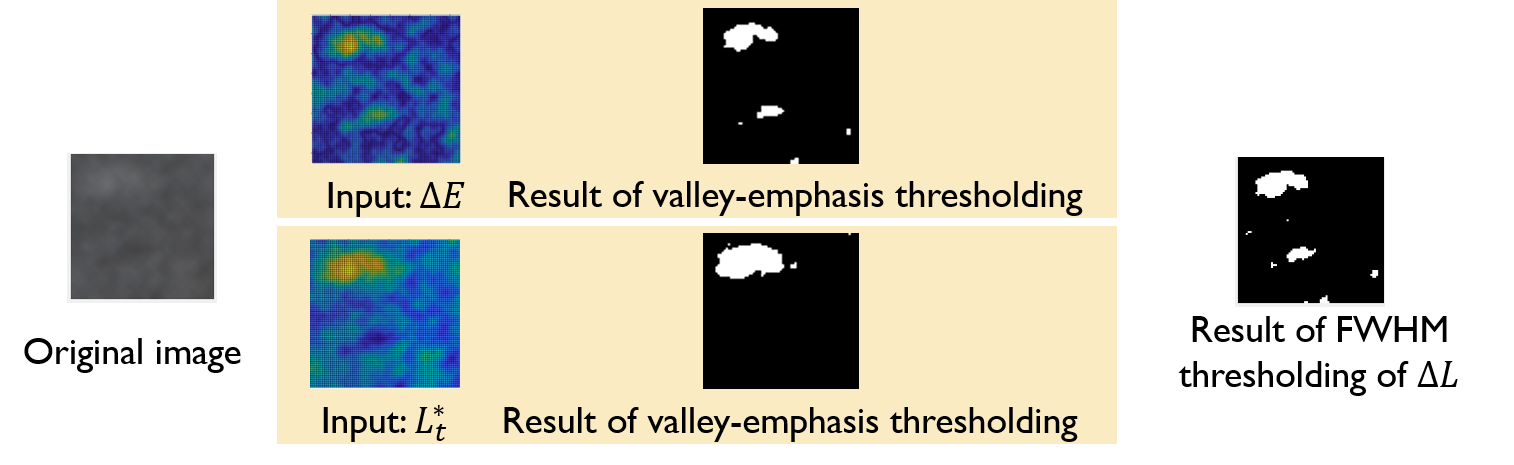}
  \caption{Results of Valley-emphasis Algorithm}
  \label{Figure:valley}
\end{figure}

In Figure \ref{Figure:valley}, we compare the results of applying the valley-emphasis algorithm followed by thresholding, to the descreened image expressed in either $\Delta E$, $L^*$ units. Figure \ref{Figure:valley} shows the results of different inputs after thresholding. The left-most image is the descreened input block, which includes a gray spot on the upper-left corner and some dispersed dark agglomerates. The result of $L^*$ seems to be more focused on the gray spot region, while the $\Delta E$ result marks out both the gray spot the and dark agglomerates. For comparison, we also show the result that is directly acquired from the FWHM (Full width at half maximum) of $\Delta L$, which is more close to the $\Delta E$ results. The reason for this difference lies in the operation to get the "$\Delta$ values", which is computing the absolute difference from average. In this manner, both dark and light regions strongly deviate from average, so that they are both marked out by the valley-emphasis algorithm. According to more comparisons on our test pages, the $L^*$ inputs tend to give more gray spot results while fewer dark spots. Depending on our actual needs, we can choose either input to maximize accuracy.

After we get the mask of defects within a block, we can conduct the analysis for its attributes:

\begin{itemize}
    \item Size (pixels): number of pixels in the defect area selected by valley-emphasis algorithm;
    \item Light / Dark: if the defect area is lighter or darker than the block average. This attribute can help us identify the type of local defect;
    \item Major and minor axis lengths: major and minor axis lengths of the equivalent eclipse of the defect area;
    \item Severity: the contrast of the defect area versus the background, defined by $\frac{\sum_{defect} \Delta E}{\sum_{background} \Delta E}$, which can also be regarded as the ratio of "defect volume" to the "background volume".
\end{itemize}

\begin{figure}[!hb]
\centering
  \includegraphics[width=\columnwidth]{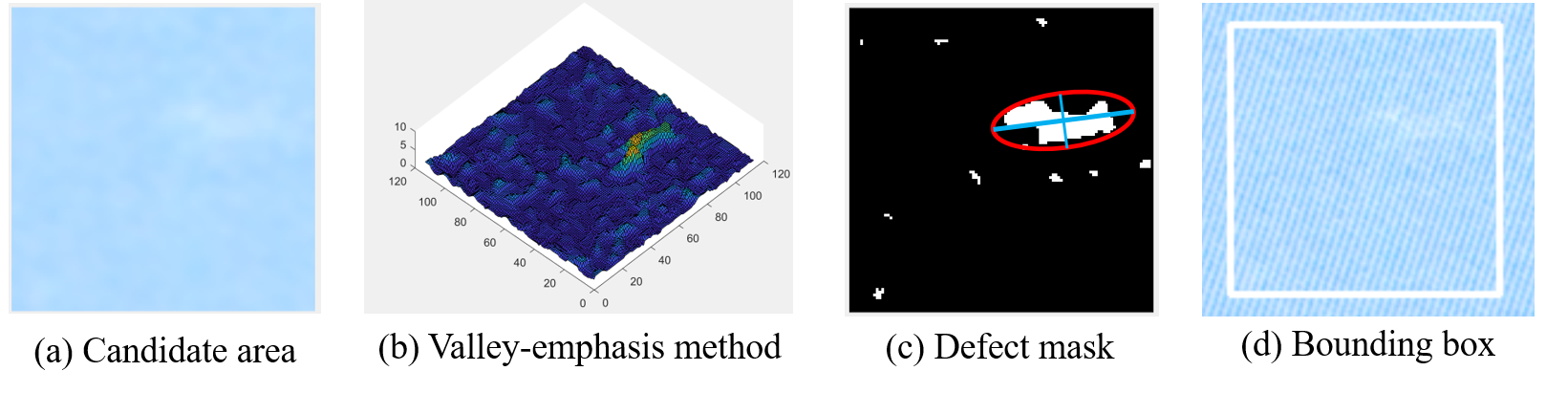}
  \caption{The processing of a candidate area in each step.}
  \label{Figure:abcd}
\end{figure}

By inspecting each ROI according to the process shown in Figure \ref{Figure:abcd}, we can get the attributes above. Along with $DDE$, $MDL$, $DDL$, these features will be used in the following steps to exclude false positives (invisible defects). To better visualize the detected visible defects, we draw a white bounding box around the block with gray spots, and a black box around the block with dark spots. We use the Connected Components algorithm to combine bounding boxes of adjacent blocks with the same type defect into one bigger box.

\begin{figure}[!hb]
  \includegraphics[width=\columnwidth]{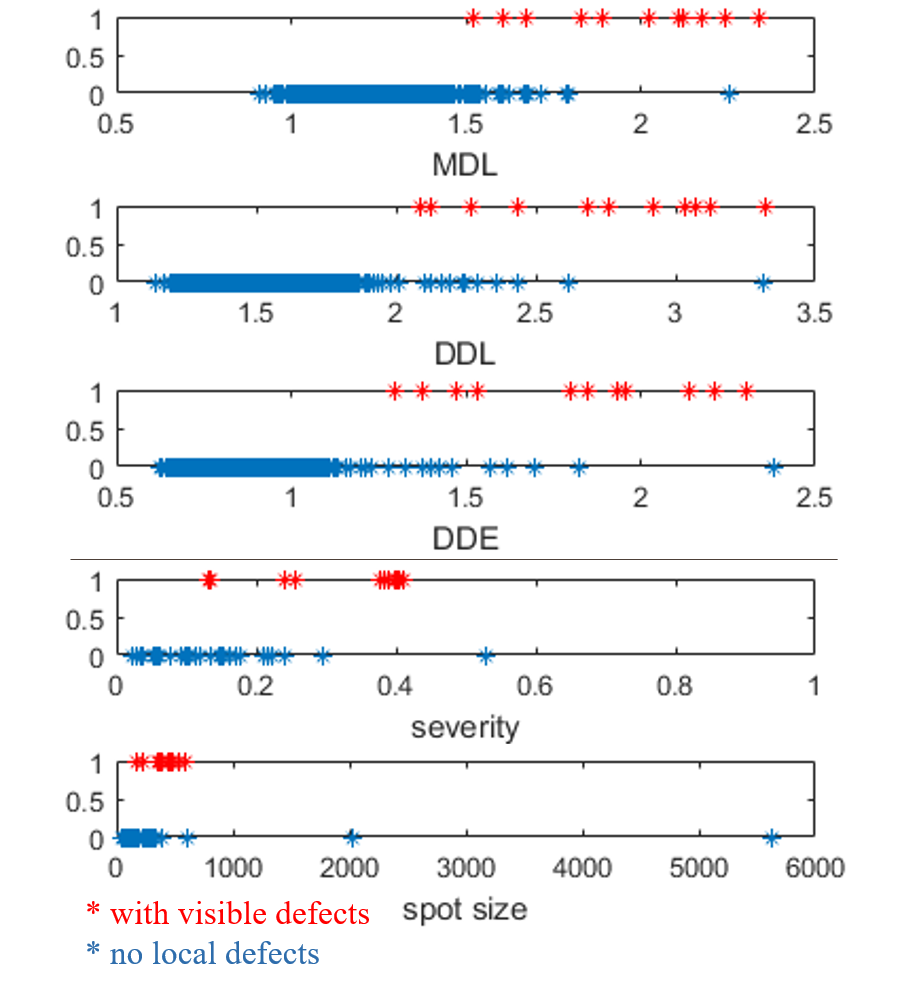}
  \caption{Distribution of each feature. The blocks with defects are marked as 1, and the other blocks as 0, so different blocks appear on the top and bottom of the graph, respectively.}
  \label{Figure:hard_thr}
\end{figure}

If we mark blocks with visible local defects as 1, and the rest as 0, we can plot their distribution versus each feature as shown in Figure \ref{Figure:hard_thr}. According to these plots, we can tell that $MDL$, $DDL$, and $DDE$ are all good metrics for choosing candidate blocks in the initial step. The severity can exclude abnormal cases (e.g. infinite values). With defect size, we can exclude cases that are too small to see. Similarly, the major and minor axis lengths can help us exclude thin lines that are imperceptible and non-local. Although all these features obviously correlate with the existence of visible local defects, there is no single threshold that divides blocks with/without defects. 

\subsubsection{Blockwise Dataset}
We create a blockwise dataset that includes several types of local defects including gray spots, pinholes, etc. for further refinement models. This dataset is from 15 test pages with 66 uniform color regions including 12 colors. These test pages are in A4 size, scanned with 600 dpi. Taking out margins and barcode areas, 67,465 blocks are sampled from all test pages, among which 1,502 blocks are marked as blocks with local defects. After initial computation, 5,043 blocks are selected as ROIs by our algorithm. 

Each block sample contains the following three types of metadata: 1) glocal features, including filename, block index, and color, which are related to the whole page; 2) blockwise features, including the block's $x$ and $y$ coordinate ranges, average $L^*$, $a^*$, $b^*$ values, $DDE$, $MDL$, $DDL$, and the ground truth of local defects. These features can help us locate the block among the raw pages, and are only related to the block itself; 3) local defect features: light/dark, defect size, equivalent eclipse's major and minor axis length, and severity. Only the 5,043 blocks that passed initial selection have the third type of features.

\begin{figure}[!hb]
\centering
  \includegraphics[width=0.75\columnwidth]{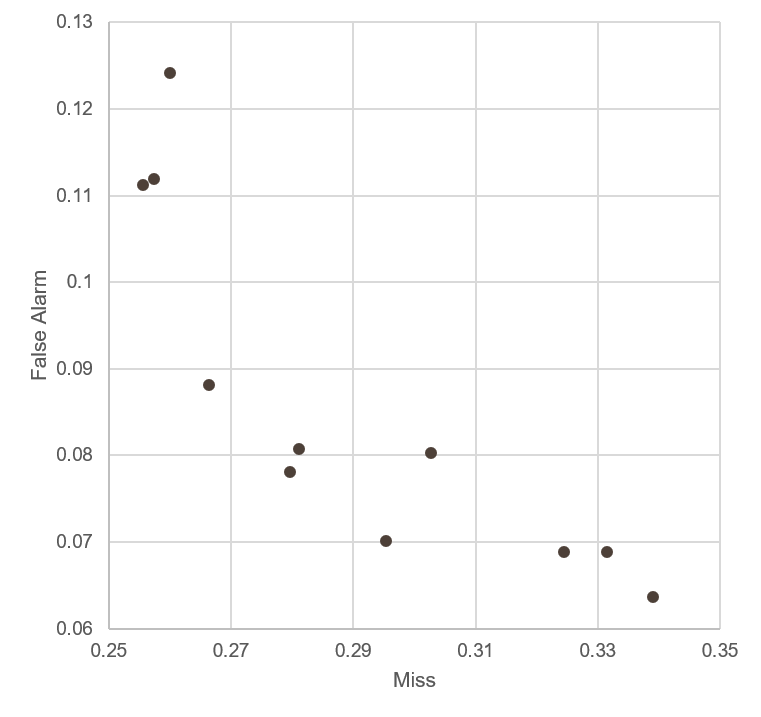}
  \caption{ROC (Receiver operating characteristic) plot}
  \label{Figure:roc}
\end{figure}

Based on our blockwise dataset, we trained a decision tree model with 7 blockwise features as decision nodes: $MDL$, $DDL$, $DDE$, defect size, major axis length, minor axis length, and severity. The selection of which feature to use and the specific split is chosen using information entropy and gain. Since in our training set, the number of blocks with defects is smaller comparing to normal blocks, we train the decision tree model with a $2\times 2$ cost matrix, where element $C(i,j)$ of this matrix is the cost of classifying an observation into class $j$ if the true class is $i$. By changing the miss-classification cost, we can get models with different miss rates and false alarms that are calculated by:

\begin{equation}
    \text{Miss Rate} = \frac{FN}{TP+FN}
\end{equation}

\begin{equation}
    \text{False Alarm} = \frac{FP}{FP+TN}
\end{equation}

where $FN$ denotes "False Negative", $TP$ is "True Positive", $FN$ is "False Negative", $FP$ is "False Positive", and $TN$ is "True Negative". The performance of our model is shown in Figure \ref{Figure:roc}, where the best result of our model is obtained when $cost = 2$, the False Alarm rate is $0.088$, while the Miss rate is $0.266$. 

\section{Results}

\begin{figure}[!hb]
\centering
  \includegraphics[width=\columnwidth]{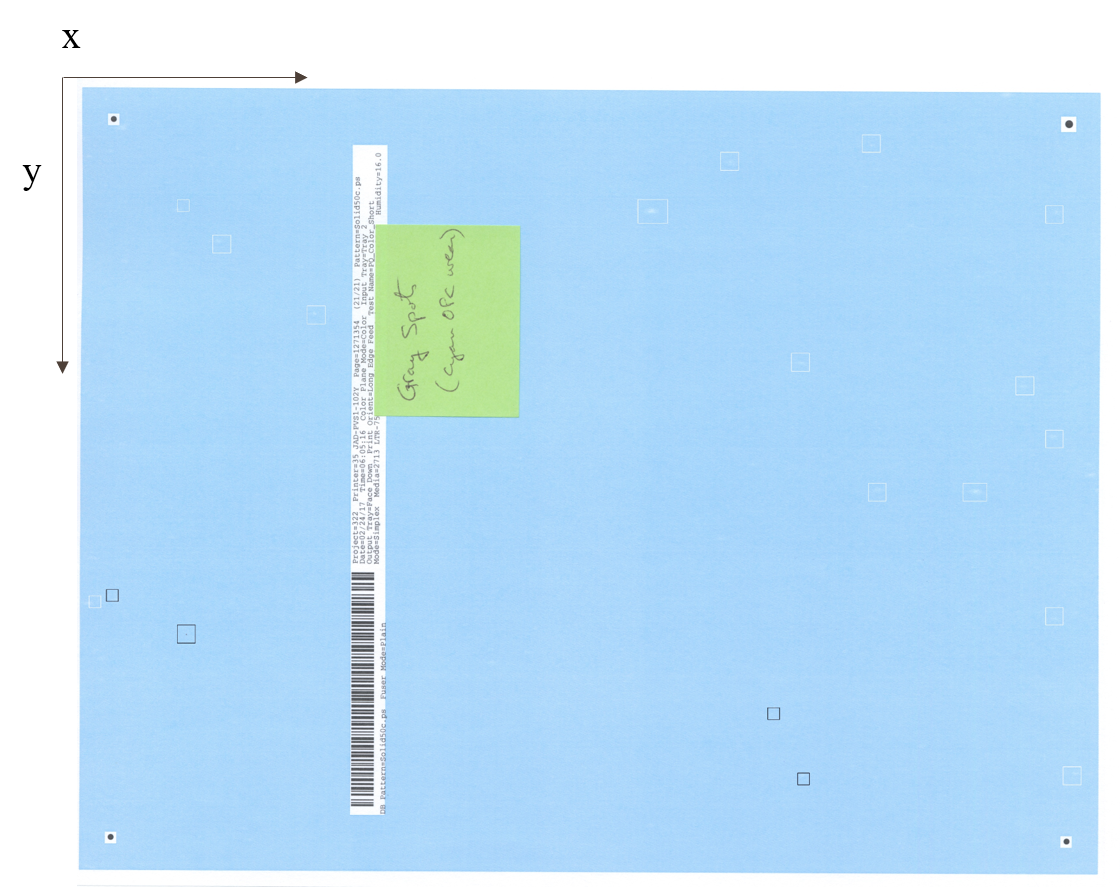}
  \caption{Detection output of a test page. The boxes indicate the defects that have been identified (White boxes stand for gray spots, black boxes represents solid spots).}
  \label{Figure:output}
\end{figure}

Taking the final output blocks of the decision tree, we can get the final detection output as shown in Figure \ref{Figure:output}. By aggregating their information we can generate a feature vector for the whole test page. There are several features that we care about: number of all types of defects on the test page, number of each type of defects, local defects' average size, and maximal and minimal size, and standard deviation of the size, average severity, maximal and minimal severity. In addition, we also want to know the average location of all local defects to determine whether or not their distribution is random. The output feature vector for a test page is listed in Table \ref{tab:fea_vec}.

\begin{table}[!ht]
\caption{Table 1: The feature vector for a test page}
\label{tab:fea_vec}
\begin{center}       
\begin{tabular}{|p{0.75\columnwidth}|p{0.15\columnwidth}|} 
\hline
Number of defects on this page & 19\\ \hline
Number of Gray Spots  & 15 \\ \hline
Number of Solid Spots & 4 \\ \hline
Average size ($mm^2$) & 0.53 \\ \hline
Max size ($mm^2$) & 1.24 \\ \hline
Min size ($mm^2$) & 0.08 \\ \hline
The standard deviation of size ($mm^2$) & 0.30 \\ \hline
Average severity & 0.16 \\ \hline
Max severity & 0.16 \\ \hline
Min severity & 0.16 \\ \hline
Average $y$ coordinate from the center of the page ($mm$) & -5.95 \\ \hline
Average $x$ coordinate from the center of the page ($mm$) & 24.27 \\ \hline
\end{tabular}
\end{center}
\end{table} 

\begin{figure}
\centering
  \includegraphics[width=\columnwidth]{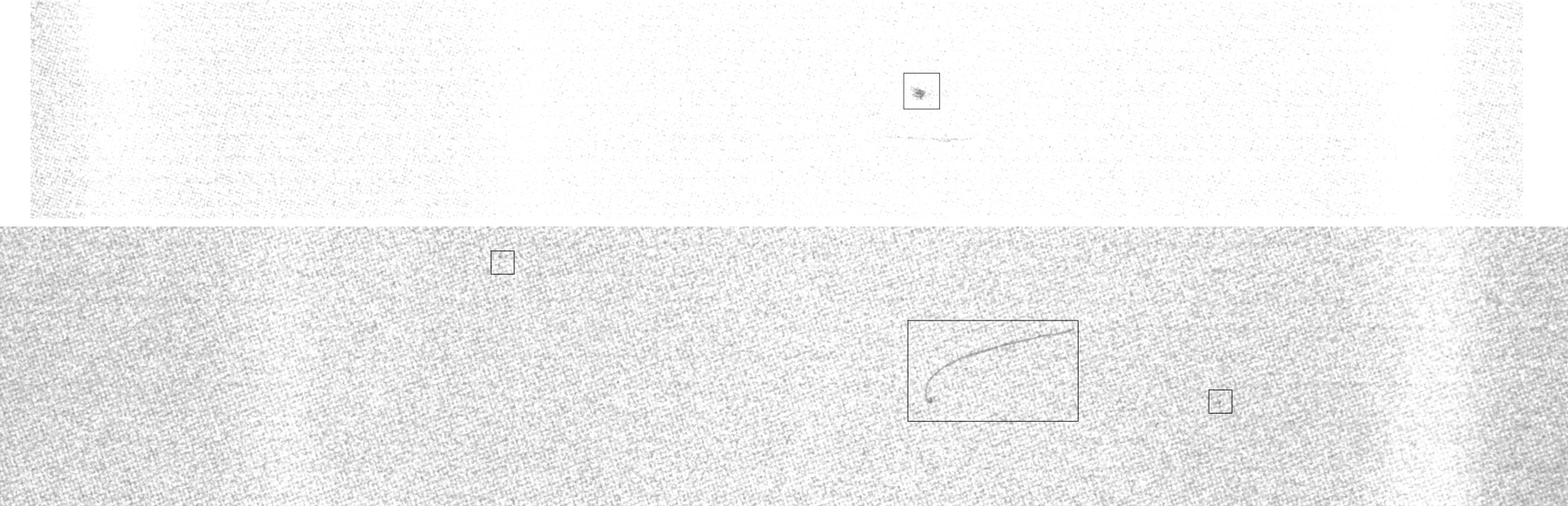}
  \caption{The detection output of touchpad products. The boxes indicate the defects that have been identified.}
  \label{Figure:pad}
\end{figure}

Our algorithm can not only be applied on print quality assessment, but also on detecting the scratches and contamination in the manufacturing of glass touchpads. Figure \ref{Figure:pad} shows that our method is robust to background noise that is generated from the matte surface and uneven lighting.

\section{Conclusion}
In this paper, we develop a coarse-to-fine method to automatically detect local defects, including the initial detection by thresholding, and the secondary refinement by a trained model. Different from previous works, we propose block-wise features to describe attributes of visible defects in the candidate area, which can help us determine the exact defect type. With these proposed features, we build a block-wise dataset of local defects for future training. A decision tree model is applied to produce more accurate results for visible local defects. Finally, we agglomerate block-wise results to generate a feature vector for each test page, which can be used for further assignment of page rank. 

\section{Acknowledgments} 
We thank Dr. Minh Q Nguyen that previously worked in our lab for providing guidance and expertise that greatly assisted the research. We are also immensely grateful to Wei Wu and Bo Jia for their support regarding the touchpad's defect detection.



\small
\bibliography{aaabib}
\bibliographystyle{plain}



\begin{biography}

\textbf{Xiaoyu Xiang} is a Ph.D. student at School of Electrical and Computer Engineering, Purdue University, supervised by Prof. Jan P. Allebach. She received her B.S. in Engineering Physics from Tsinghua University (2011), Beijing, China. Her research interests include deep learning, image quality assessment and enhancement, face alignment and recognition.

\textbf{Renee Jessome} received a B.S. in Mechanical Engineering from the Rochester Institute of Technology, Rochester, NY, in 1992. She has worked on a variety of printing devices over the past 14 years from the smallest LaserJet printers to Indigo Printing presses. She currently leads a Product Development team focused on high-end LaserJet Printers and MultiFunction Printers at HP Inc. in Boise, Idaho. During her early career at HP, she was developing storage devices such as disk drives, tape drives, and disk arrays. 

\textbf{Eric Maggard} received his B.S. degree in Physics from Northwest Nazarene University, Nampa, Idaho in 1991 and the M.S. degree in Computer Science specializing in image analysis and processing from Walden University in 2006. He is an Expert Imaging Scientist in the LaserJet Hardware Division and has developed programs and image quality algorithms for the last 15 years that are used in the testing of LaserJet print and scan image quality. His interests include machine vision, object recognition, machine learning, robot control, and navigation. 

\textbf{Yousun Bang} is a manager of Image Quality Part in Imaging Lab at HP Printing Korea Co. Ltd.  She received her B.S. and M.S. in mathematics from Ewha Womans University, Seoul, Korea in 1994 and 1996, and her Ph.D. in the school of Electrical and Computer Engineering, Purdue University, West Lafayette, Indiana in 2005. She worked for Samsung Advanced Institute of Technology and Samsung Electronics Company from 2004 to 2017. Her current research interests include image quality diagnosis and metrics and ML/DL based prediction for smart device services.

\textbf{Minki Cho} is the engineer of HP Printing Korea. He received B.S.(1997) and M.S.(1999) in electrical engineering from Inha University, Korea. From 2003 to 2017, he worked for Samsung Electronics and Samsung Advanced Institute of Technology. His research areas are print image processing, print image quality diagnosis \& calibration. Recently, he is researching above interesting using ML \& DL.

\textbf{Jan P. Allebach} is Hewlett-Packard Distinguished Professor of Electrical and Computer Engineering at Purdue University. Allebach is a Fellow of the IEEE, the National Academy of Inventors, the Society for Imaging Science and Technology (IS\&T), and SPIE. He was named Electronic Imaging Scientist of the Year by IS\&T and SPIE and was named Honorary Member of IS\&T, the highest award that IS\&T bestows. He has received the IEEE Daniel E. Noble Award and is a member of the National Academy of Engineering. He currently serves as an IEEE Signal Processing Society Distinguished Lecturer (2016-2017).

\end{biography}

\end{document}